\pdfoutput=1

\documentclass[11pt]{article}

\usepackage[]{latex/acl}

\usepackage{times}
\usepackage{latexsym}

\usepackage[T1]{fontenc}
\usepackage[normalem]{ulem}
\usepackage{algorithm}
\usepackage{multirow}
\usepackage{amsmath} 
\usepackage{enumitem}

\usepackage{amsfonts}

\usepackage{amsmath}
\usepackage{booktabs}

\usepackage{subcaption}
\usepackage{algorithm}
\usepackage{algpseudocode}
\usepackage[most]{tcolorbox}
\usepackage[utf8]{inputenc}
\usepackage[section]{placeins}
\usepackage[utf8]{inputenc}

\usepackage{microtype}

\usepackage{inconsolata}

\title{Accelerating Adaptive Retrieval Augmented Generation via Instruction-Driven Representation Reduction of Retrieval Overlaps}

\author{\\
Jie Ou$^1$, Jinyu Guo$^{1,*}$, Shuaihong Jiang$^1$, Zhaokun Wang$^1$,\\
Libo Qin$^2$, Shunyu Yao$^3$\and Wenhong Tian$^1$ \\
        $^1$ School of Information and Software Engineering, \\
        University of Electronic Science and Technology of China\\
        $^2$ Central South University\\
        $^3$ Big data and artificial intelligent institute, China Telecom Research Institute\\
        * Corresponding Author: Jinyu Guo, Email: guojinyu@uestc.edu.cn\\
        }

\begin{document}
\maketitle
\begin{abstract}
Retrieval-augmented generation (RAG) has emerged as a pivotal method for expanding the knowledge of large language models. To handle complex queries more effectively, researchers developed Adaptive-RAG (A-RAG) to enhance the generated quality through multiple interactions with external knowledge bases. Despite its effectiveness, A-RAG exacerbates the pre-existing efficiency challenges inherent in RAG, which are attributable to its reliance on multiple iterations of generation. Existing A-RAG approaches process all retrieved contents from scratch. However, they ignore the situation where there is a significant overlap in the content of the retrieval results across rounds. The overlapping content is redundantly represented, which leads to a large proportion of repeated computations, thus affecting the overall efficiency. To address this issue, this paper introduces a model-agnostic approach that can be generally applied to A-RAG methods, which is dedicated to reducing the redundant representation process caused by the overlapping of retrieval results. Specifically, we use cache access and parallel generation to speed up the prefilling and decoding stages respectively. Additionally, we also propose an instruction-driven module to further guide the model to more effectively attend to each part of the content in a more suitable way for LLMs. Experiments show that our approach achieves 2.79 and 2.33 times significant acceleration on average for prefilling and decoding respectively while maintaining equal generation quality.
\end{abstract}

\section{Introduction}
\textit{Work smarter, not harder.}
\vspace{3pt}
\hrule
\vspace{3pt}
\hfill\textit{Allan F. Mogensen, 1930s}
\newline

Large Language Models (LLMs) (e.g., LLaMA-2, PaLM, and GPT-4 \cite{touvron2023llama,anil2023palm,achiam2023gpt}) have demonstrated remarkable capabilities across various natural language processing tasks. To address the growing demand for knowledge-intensive and factually grounded responses, Retrieval-Augmented Generation (RAG) has emerged as a promising paradigm to mitigate the inherent knowledge limitations of LLMs \cite{lewis2020retrieval,borgeaud2022improving,ram2023context,jiang2023active,asaiself,gao2023retrieval,gupta2024comprehensive}. It enhances the performance by retrieving relevant information from external knowledge base to generate contextually appropriate and evidence-based responses.
\begin{figure}[t]
    \centering
    \includegraphics[width=0.9\linewidth]{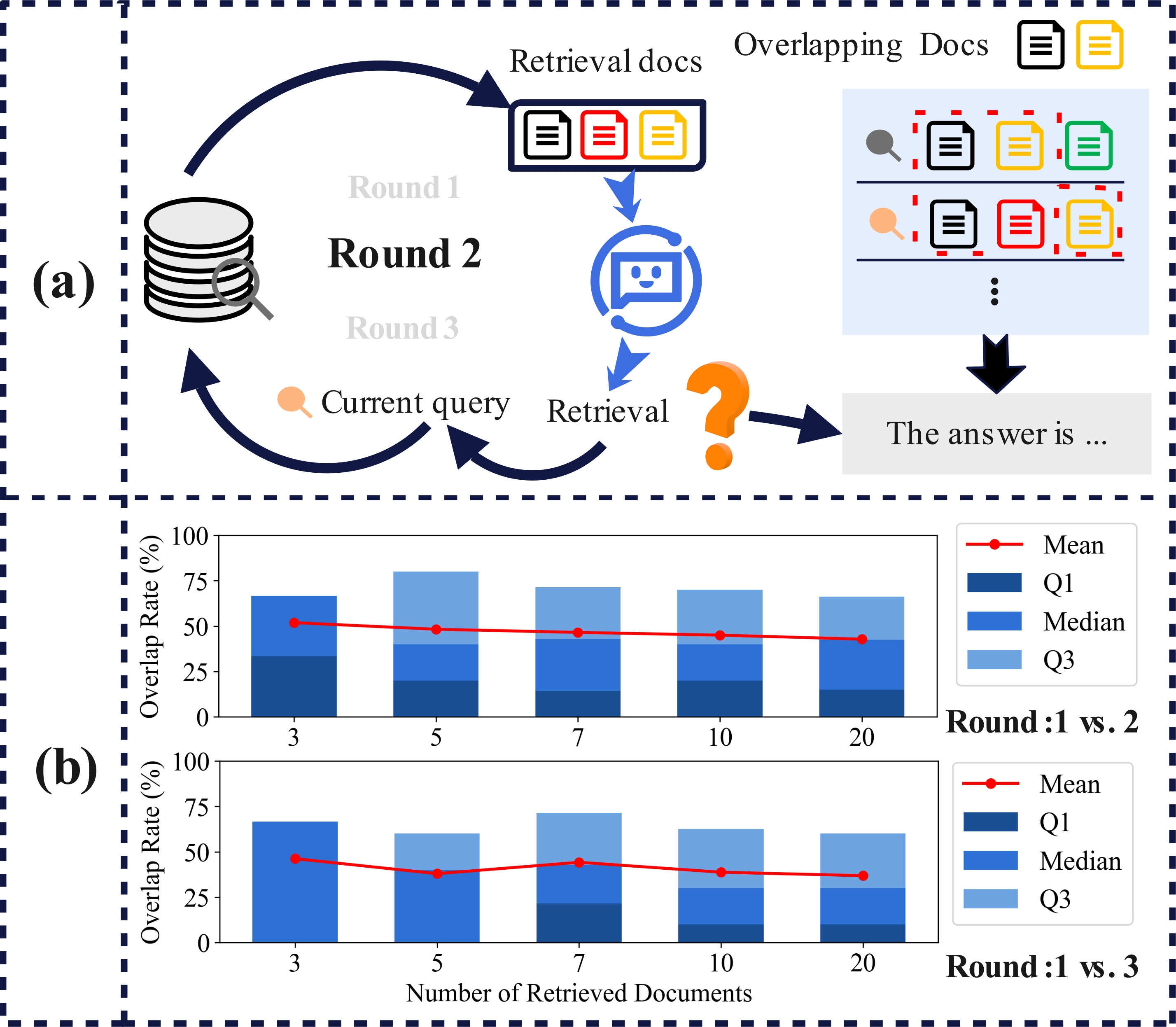}
    \caption{(a) The pipeline of A-RAG. (b) Analysis of document overlap between first and later retrievals (rounds 2-3) using 1000 2WikiMultihopQA samples.}
    \label{fig:1}
\end{figure}

Conventional RAG interacts with the external knowledge bases once and combines the retrieved documents with the query as input to LLMs, this can also be called single-round RAG. Nevertheless, single-round RAG encounters challenges in handling complex user inquiries, as it is difficult to extract sufficient information by a single interaction. 
Consequently, an increasing number of researchers have begun to focus on and propose Adaptive-RAG (A-RAG) \cite{borgeaud2022improving, ram2023context,mallen2023not,asaiself,trivedi2023interleaving,jiang2023active,zhao2023verify,yang2023improving,su2024dragin,zhang2024retrievalqa, lichain,yao2024seakr,jeong2024adaptive,wang2024retriever}. Figure 1(a) illustrates the process of A-RAG, which dynamically determines whether to continue retrieval and adjusts the retrieval content based on the quality of the generated responses. Through iterative interactions with external knowledge bases, A-RAG obtains more comprehensive and valuable information to generate accurate and comprehensive answers.

Although A-RAG enhances the performance, it further exacerbates the inherent efficiency problems of the RAG due to external interactions and increased computation. Within its post-retrieval generation phase, existing methods process all retrieved contents from scratch. However, they overlooked a situation where, during a single A-RAG process, the high similarity among multiple rounds of queries results in significant overlap in the content of the retrieved results across these rounds, especially between adjacent rounds. Figure \ref{fig:1}(b) shows the overlap ratio of documents between rounds with different settings for the number of retrieved documents. These overlapping contents are redundantly represented in each round, which leads to a large proportion of repeated computations, thereby affecting the overall efficiency.

To tackle this issue, we propose \textbf{I}nstruction-\textbf{D}riven \textbf{R}epresentaion \textbf{R}eduction (\textbf{IDR$_2$}), a model-agnostic approach widely applicable to the A-RAG methods. It aims to improve the efficiency of A-RAG by eliminating repetitive representations of overlapping content and redundant autoregressive representation rounds. Concretely, we leverage representation reduction techniques in both the prefilling and decoding processes of the generation phase. During prefilling, we propose the Cross-Iteration Cache Sharing (CICS) module, which establishes a shared memory repository for document representations. It enables subsequent processing iterations to bypass repetitive computations through cached intermediate results, thereby reducing computational overhead for overlapping content. In addition, to further direct the model to more effectively attend to the various parts of the content after prefilling, we introduce the Instruction-driven Deduplication Guidance Reinforcement (IDGR) module. 
It leverages the instruction-following capabilities of LLMs to implement context-aware filtration, prioritizing semantically relevant cached information while suppressing redundant content through explicit linguistic guidance in a more suitable way for LLMs.
In the decoding process, we propose a novel Information-Guided Parallel Generation (IGPG) module, which leverages the correlation between retrieved documents and the generated results. By integrating phrasal fragments as inputs at each autoregressive step, IGPG enables parallel generation. It reduces the autoregressive representation rounds to achieve acceleration.

We extensively evaluate our proposed method on multiple datasets. The experimental results show that our IDR$_2$ significantly reduces the representation process, which achieves 2.79 and 2.33 times acceleration for prefilling and decoding, consequently accelerating the entire A-RAG workflow by 2.0 times on average across various A-RAG approaches while maintaining the performance of generation.
In summary, our contributions are mainly three-fold:
\begin{itemize}
\item [1.] We propose IDR$_2$, an acceleration approach for A-RAG based on representation reduction techniques. It eliminates repetitive representations during the prefilling process and reduces autoregressive representation redundancy in the decoding phase, thereby speeding up the entire A-RAG workflow almost without performance loss. 

\item [2.] We develop the Instruction-driven Deduplication Guidance Reinforcement module, which leverages the instruction-following capabilities of LLMs to further direct the model to more effectively attend to the various parts of the content after prefilling in a more suitable way for LLMs.

\item [3.] Experimental results demonstrate that our approach significantly enhances the efficiency of various A-RAG methods while exhibiting robust adaptability across diverse scenarios and various LLM scales.
\end{itemize}

\section{Related Works}
\subsection{Adaptive-RAG}
Generating satisfactory answers by single-round RAG remains challenging \cite{komeili2021internet,zhu2021retrieving,liu2024retrieval,jiang2023active, yang2023improving,ni2024llms}.
A-RAG was developed to address this limitation by enabling iterative interactions with external knowledge bases, leading to more accurate and complete answers through multiple retrieval rounds \cite{jiang2023active, yang2023improving,ni2024llms}.
Current A-RAG approaches can be classified into two categories: 1) Rule-based strategies, such as per-sentence iteration \cite{trivedi2023interleaving}, sliding window tokens \cite{borgeaud2022improving, ram2023context}, and contextual learning \cite{zhao2023verify, zhang2024retrievalqa, lichain}. 2) Self-perception strategies evaluate output confidence through internal states \cite{yao2024seakr}, LLM output layer \cite{jiang2023active,yang2023improving,su2024dragin}, or explicit language-level feedback \cite{asaiself}.
In contrast to existing works that primarily focus on performance enhancement, to the best of our knowledge, we are the first to investigate the multi-turn document overlapping problem in A-RAG and improve efficiency through its resolution.

\subsection{Inference Acceleration}
\textbf{Efficiency of LLM.} Research on improving LLM efficiency can generally be classified into two categories: traditional optimization techniques (such as quantization, pruning, knowledge distillation, etc.) and LLM-specific optimizations which include: 1) Early Exiting \cite{teerapittayanon2016branchynet,xin2020deebert,zhou2020bert,kong2022accelerating,yangpredictive,bae2023fast}, uses prediction heads at different layers to enable early token exit when confidence thresholds are met.
2) Token Pruning \cite{goyal2020power,kim2021length,wang2021spatten,kim2022learned,hou2022token,zhang2023h2o}, retains only critical tokens based on importance ranking.
3) Speculative Decoding \cite{chen2023accelerating,leviathan2023fast,spectoraccelerating,yang2023inference,zhang2023draft,kim2024speculative,miao2024specinfer}, uses efficient smaller models to generate candidate tokens for batch verification by LLM.

\textbf{Efficiency of RAG.} Speculative retrieval \cite{zhangaccelerating} reduces retrieval overhead through local retrieval and verification mechanisms. Parallel document processing \cite{merthsuperposition} mitigates attention complexity by processing multiple documents independently rather than concatenating them. Document preprocessing \cite{lu2024turborag} improves prefilling efficiency by preprocessing and storing document representation for the whole knowledge base. Small expert LMs generate initial document-specific drafts for LLM verification \cite{wang2024speculative}.
In contrast to prior work, our IDR$_2$ is a general framework. 
It enhances A-RAG efficiency by representation reduction with instruction-guided information extraction.

\section{Methods}
\begin{figure*}[h]
    \centering
    \includegraphics[width=0.95\linewidth]{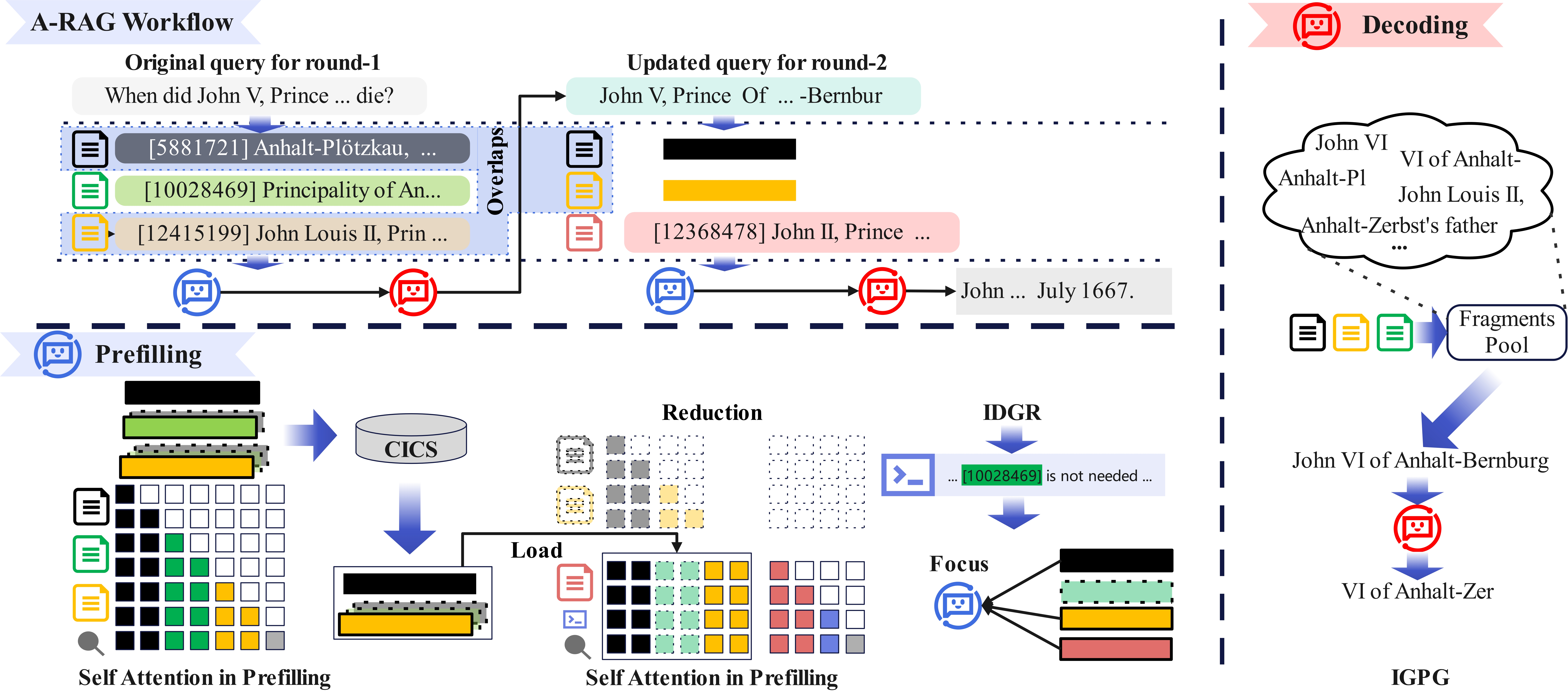}
    \caption{The pipeline of our IDR$_2$. the same color indicates the same document and representation. 
    }
    \label{fig:2}
\end{figure*}

Figure \ref{fig:2} details the workflow of IDR$_2$, and we divide the internal process of each iteration round in A-RAG into three phases: retrieval, prefilling, and decoding. When processing a query, cached document representations in the CICS module are first verified. These representations are loaded and combined with uncached documents for prefilling, where the IDGR module guides LLMs to prioritize relevant content while filtering noise. Before each autoregressive step, matching subsequent phrase fragments are queried in the IGPG module to enable parallel generation. Ultimately, these approaches enhance the overall efficiency of A-RAG. Details of each module are given respectively in the remainder of this section.

\subsection{Cross-Iterative Cache Sharing (CICS)}
A-RAG \cite{zhao2023verify,jiang2023active,yang2023improving,su2024dragin,zhang2024retrievalqa} relies on multiple rounds of retrieval-generation interactions to generate answers. We observe substantial document overlap between adjacent retrieval rounds. In light of this, duplicate representation for overlapping documents introduces computational redundancy, hampering the efficiency of A-RAG.

In order to avoid duplicate representation for overlapping documents, we propose the cross-iterative cache sharing (CICS) module.
CICS initializes a cache space $\mathbb{C}$ to store Key-Value pairs corresponding to the documents retrieved in each round for each query $q$. 
Because A-RAG may modify the query at the end of each iteration, we denote the initial user query as $q_0$. 
At round $t$, the retrieval operation $D_t=\textbf{Retrieve}(q_t)$ returns a set of $n$ documents $D_t=\{d_1^t,d_2^t,...,d_n^t\}$ from the external knowledge base. $n$ denotes the maximum number of documents that can be retrieved in each round.
At round $t$, the LLM generates a sequence of tokens $A_t=\{a_t^1, a_t^2, ..., a_t^m\}$ by:
\begin{equation} 
\label{eq:2} 
a_t^1, K_t, V_t = \textbf{LLM}_P(q_t, D_t,A_{<t}) 
\end{equation} 
\begin{equation} 
\label{eq:3} 
\begin{aligned} a_t^i, k_t^i, v_t^i &= \textbf{LLM}_D(q_t, D_t, A_{<t}, a_t^{<i}, K_t, V_t)\\ 
K_t &= \textbf{Concat}(K_t, k_t^i)\\ 
V_t &= \textbf{Concat}(V_t, v_t^i)
\end{aligned}
\end{equation}
where the inference process of LLM is divided into two stages: prefilling and decoding, represented with $\textbf{LLM}_P$ and $\textbf{LLM}_D$ respectively. The $K_t$ and $V_t$ represent the Key-Value pairs for $D_t$ and $A_t$. The $a_t^i$ denotes the $i^{th}$ generated token and $m$ represents the number of generated tokens. 
The Eq.(\ref{eq:3}) denoted the autoregression process, which can only generate one token at each step. It needs to be executed through multiple steps to obtain the complete $A_t$. CICS stores the $K_t, V_t$ from Eq.(\ref{eq:2}), which is the representation for $D_t$ as shown in Figure \ref{fig:2}.

After receiving retrieved results $D_t$, CICS extracts existing representation from the cache to avoid duplicate representation. Specifically, as the example shown in Figure \ref{fig:2}, CICS directly loads the representation of documents \#5881721 and \#12415199, then integrates them with the text of document \#12368478 for generation. This process can be formalized as:
\begin{equation}
\label{eq:4}
\begin{aligned}
K_t^o,V_t^o, D_t^o &= \textbf{Filter}(D_t, \mathbb{C})\\
a_t^1, K_t, V_t &= \textbf{LLM}_P(q_t, D_t^n,A_{<t}, K_t^o, V_t^o)
\end{aligned}
\end{equation}
where $D_t^o$ is the document set that has been processed in the previous round and appears again at the current round $D_t$. We can directly obtain its corresponding representation $K_t^o$ and $V_t^o$ from $\mathbb{C}$ without re-processing. For the new document set $D_t^n = D_t \setminus D_t^o$ at the current round, the prefilling is still required. 

Eq.(\ref{eq:4}) optimizes the prefilling phase by reusing the cached representation of overlapping documents in CICS, avoiding redundant computation.

\subsection{Instruction-driven Deduplication Guidance Reinforcement (IDGR)}
The CICS provides efficient representation reuse across rounds. Notably, the Key-Value representation of each document $d_i^t$ incorporates information from previously processed documents through self-attention. As an example illustrated in Figure \ref{fig:2}, at round 1, the representation of document \#12415199 contains information from documents \#5881721 and \#10028469 due to the self-attention mechanism.

The IDGR employs natural language instructions $I_t$ in the prompt to guide the LLM in filtering redundant cached information. This module helps the model focus on relevant content for the current round, thereby ensuring high-quality generation. At each iteration, the instruction $I_t$ is automatically generated based on two rules and is subsequently utilized during the prefilling phase, as:
\begin{equation}
\label{eq:6}
a_t^1, K_t, V_t = \textbf{LLM}_P(q_t, D_t^n,A_{<t}, R_t^o, I_t)
\end{equation}
where $R_t^o$ is the $K_t^o, V_t^o$.
As shown in Figure \ref{fig:2}, the representation of document \#10028469 is excluded from the computation. Furthermore, explicit instructions guide the LLM to ignore the information contained in the representation of document \#12415199.

The instruction $I_t$ is generated according to two rules: First, we use document identifiers to explicitly tell the LLM which documents are relevant and which are irrelevant for the current generation. This helps the LLM focus on pertinent context while filtering out redundant information. Second, we provide document relevance rankings to the LLM. In cases where we cannot directly adjust the input order of documents based on their importance, explicit instructions guide the LLM to prioritize more relevant documents.

In practice, $I_t$ can be implemented with simple natural language directives, leveraging the strong semantic understanding capabilities of LLMs. For instance, in the second round shown in Figure \ref{fig:2}, the instruction is formulated as: ``\textit{\#5881721 ... are related docs. \#10028469 is unrelated. The relevance scores are ...}''. The relevance scores can be obtained through either retriever rankings\footnote{https://github.com/Muennighoff/sgpt} or model-based scoring methods\footnote{https://huggingface.co/BAAI/bge-reranker-base}. While scoring methods are not the focus of our study, they can be seamlessly integrated into our approach.

\subsection{Information-Guided Parallel Generation (IGPG)}
\begin{figure}[h]
    \centering
    \includegraphics[width=0.95\linewidth]{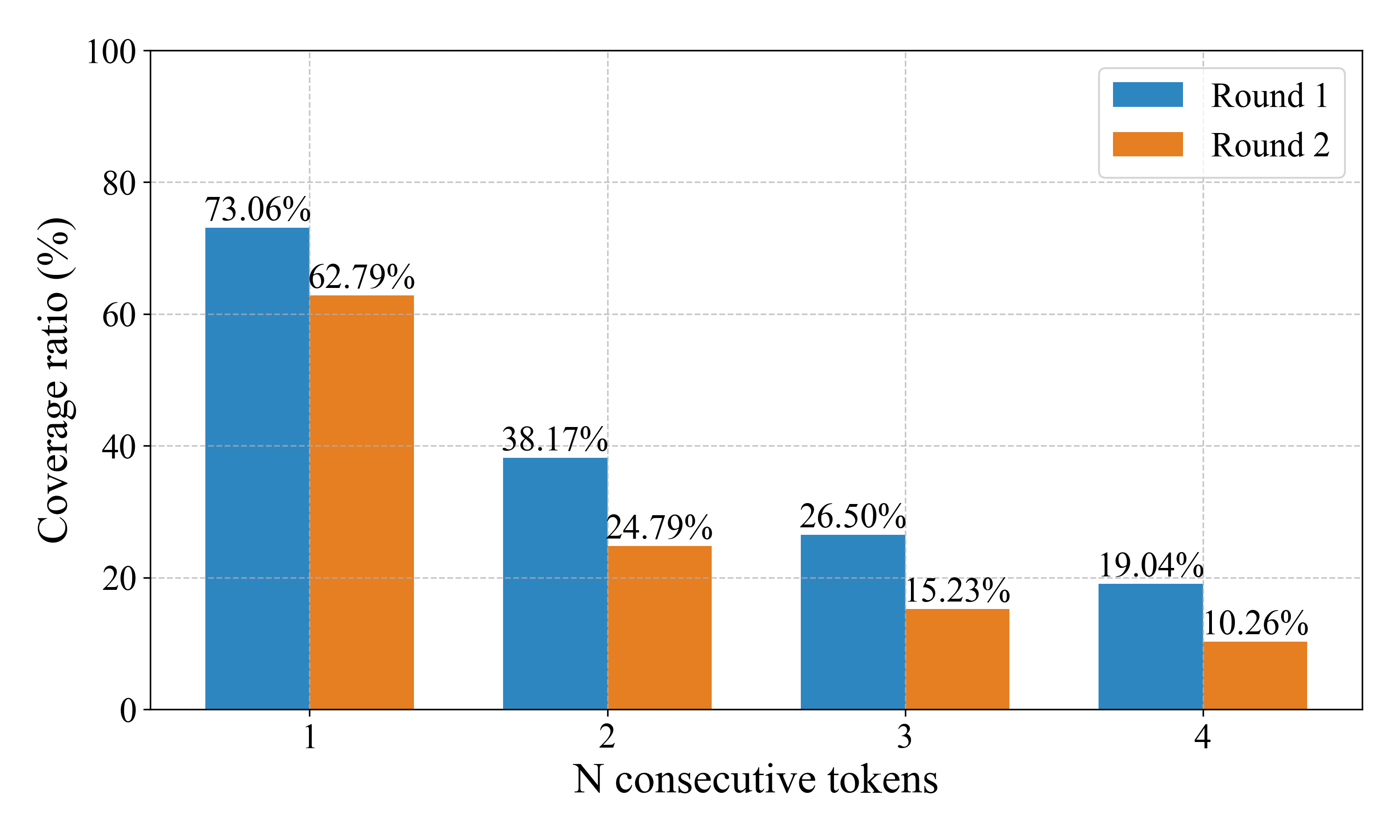}
    \caption{The x-axis represents the length of consecutive token combinations in the generated results. The y-axis represents the proportion of all combinations in the generated results that appear in the retrieved results. (LLaMA2-7B, 2WikiMultihopQA). }
    \label{fig:coverage}
\end{figure}

RAG differs from standard LLM applications in its access to extensive external context. For a given query $q$, standard LLM inference can only utilize $q$ itself, whereas RAG leverages both $q$ and retrieved documents $D_t$. Furthermore, in different iterations of A-RAG, the LLM-generated content maintains high relevance to the retrieved documents $D_t$, as illustrated in Figure \ref{fig:coverage}. This relevance plays a crucial role in determining the current round's output. The LLM generates a significant amount of existing content. The autoregressive process of token-by-token generation primarily contributes to the low efficiency. In contrast to standard LLM generation, we have the information of $D_t$, which contains a substantial portion of the content that LLM must generate. Consequently, we can utilize the information from $D_t$ to guide LLM in avoiding the independent autoregressive representation process for existing content. By constructing phrase fragments and verifying multiple tokens during autoregression, parallel generation can be achieved.

Different with traditional speculative decoding approaches, IGPG requires neither the construction of small language models nor training. During each RAG iteration, IGPG uses $D_t$ to construct a approximate probabilistic language model $P(x_t|x_1,...,x_{t-1}) \approx P(x_t|x_{t-N+1},...,x_{t-1})$, $N$ is the number of recent tokens. 

Our IGPG has two steps as speculative decoding approaches: draft generation and LLM parallel generation. In the draft generation, by iterating $M$ steps, the module constructs a $M$-length draft token sequence $\hat{A}_{t,k}=\{\hat{a}_{t,k}^1, \hat{a}_{t,k}^2, ..., \hat{a}_{t,k}^M\}$, $k$ denotes the $k$-th autoregression step. During LLM parallel generation, the LLM validates the draft token sequence through a forward pass by $P(\bar{a}_{t,k}^{i+M+1}|a_{t,k-1}^{\leq i}, \hat{a}_{t,k}^{1}, ..., \hat{a}_{t,k}^{M})$. The LLM examines whether each draft token aligns with the current output, as Eq.(\ref{eq:last}). When a draft token $\hat{a}_{t,k}^j$ fails validation, it is substituted with the LLM's prediction $\bar{a}_{t,k}^{j-1}$, and draft generation resumes from this point until the complete sequence $A_t$ is produced. 
\begin{equation}
\label{eq:last}
a_{t,k}^{j} = \begin{cases}
\hat{a}_t^j, & \text{if } \hat{a}_{t,k}^j = \bar{a}_{t,k}^{j-1}, j \geq 1\\
\bar{a}_{t,k}^{j-1} & \text{otherwise \& stop}
\end{cases}
\end{equation}
where $\{\bar{a}_{t,k}^{0}, ..., \bar{a}_{t,k}^{i+M+1}\} = \textbf{LLM}_{D}(a_{t,k-1}^{\leq i}, \hat{A}_{t,k})$, $\bar{a}_{t,k}^{0}$ corresponds to $a_{t,k-1}^{i}$ autoregressive.

If at least one token in the $\hat{A}_{t,k}$ is validated by the LLM, combined with the token generated by the LLM itself, the total number of generated tokens is at least two, achieving a least 2$\times$ speedup.
\section{Experiments}
\subsection{Language Models}
We employed the LLaMA2-7B/13B \cite{touvron2023llama} (L2-7B/13B) and the Vicuna-7B/13B \cite{chiang2023vicuna} (V-7B/13B). The Vicuna model is a chatbot fine-tuned on ShareGPT conversation data. It contains different knowledge from LLaMA and has an impact on A-RAG inference. 

\begin{table*}[h]
\resizebox{\textwidth}{!}{
\centering
\begin{tabular}{c|c|ccc|ccc|ccc|ccc}
\toprule
\multicolumn{1}{c|}{\multirow{2}{*}{Methods}} & \multirow{2}{*}{LLMs} & \multicolumn{3}{c|}{2WikiMultihopQA} &\multicolumn{3}{c|}{HotpotQA} &\multicolumn{3}{c|}{StrategyQA} &\multicolumn{3}{c}{IIRC} \\ 
\cline{3-14}
\multicolumn{1}{c|}{} & & \multicolumn{1}{c}{Pref.} $\uparrow$ & \multicolumn{1}{c}{Deco.} $\uparrow$ & E2E $\uparrow$ &\multicolumn{1}{c}{Pref.} $\uparrow$ & \multicolumn{1}{c}{Deco.} $\uparrow$ & E2E $\uparrow$&\multicolumn{1}{c}{Pref.} $\uparrow$ & \multicolumn{1}{c}{Deco.} $\uparrow$ & E2E $\uparrow$&\multicolumn{1}{c}{Pref.} $\uparrow$ & \multicolumn{1}{c}{Deco.} $\uparrow$ & E2E $\uparrow$\\
\midrule

\multicolumn{1}{c|}{\multirow{3}{*}{FLRAG+IDR$_2$}}& L2-7B & \underline{2.23}$\times$ & \underline{2.37}$\times$ &\underline{1.75}$\times$ &2.27$\times$&1.85$\times$&1.53$\times$ &1.75$\times$&1.49$\times$&1.40$\times$ &\underline{3.08}$\times$&\underline{2.76}$\times$&\underline{2.10}$\times$ \\

\multicolumn{1}{c|}{} & L2-13B & 2.12$\times$ & 2.36$\times$ &1.64$\times$ &\underline{2.29}$\times$&\underline{2.00}$\times$&\underline{1.54}$\times$ &\underline{1.76}$\times$&\underline{1.62}$\times$&1.40$\times$ &2.70$\times$&2.65$\times$&1.83$\times$ \\

\multicolumn{1}{c|}{} & V-13B & 2.05$\times$ & 2.25$\times$ &1.60$\times$ &2.22$\times$&1.97$\times$&1.51$\times$ &1.65$\times$&1.55$\times$&1.31$\times$ &2.66$\times$&2.72$\times$&1.82$\times$ \\
\midrule

\multicolumn{1}{c|}{\multirow{3}{*}{FSRAG+IDR$_2$}}& L2-7B & 2.04$\times$ & \underline{2.64}$\times$ &\underline{2.32}$\times$ &\underline{2.41}$\times$&1.88$\times$&\underline{1.69}$\times$ &\underline{2.36}$\times$&\underline{1.97}$\times$&\underline{1.79}$\times$ &\underline{2.92}$\times$&\underline{2.88}$\times$&\underline{2.20}$\times$ \\

\multicolumn{1}{c|}{}& L2-13B & \underline{2.11}$\times$ & 2.47$\times$ &1.79$\times$ &2.16$\times$&\underline{2.05}$\times$&1.59$\times$ &1.74$\times$&1.59$\times$&1.44$\times$ &2.16$\times$&2.65$\times$&1.74$\times$ \\

\multicolumn{1}{c|}{} & V-13B & 2.23$\times$ & 2.31$\times$ &1.75$\times$ &2.15$\times$&2.31$\times$&1.66$\times$ &2.02$\times$&1.78$\times$&1.53$\times$ &2.35$\times$&2.69$\times$&1.92$\times$ \\
\midrule

\multicolumn{1}{c|}{\multirow{3}{*}{FLARE+IDR$_2$}}& L2-7B & \underline{2.81}$\times$ & \underline{2.74}$\times$ &\underline{2.60}$\times$ &\underline{3.12}$\times$&\underline{\textbf{2.61}}$\times$&\underline{\textbf{2.31}}$\times$ &\underline{2.71}$\times$&\underline{\textbf{2.27}}$\times$&\underline{1.77}$\times$ &3.54$\times$&2.97$\times$&2.89$\times$ \\

\multicolumn{1}{c|}{}& L2-13B & 1.98$\times$ & 2.24$\times$ &2.16$\times$ &2.42$\times$&1.87$\times$&1.80$\times$ &2.45$\times$&1.71$\times$&1.69$\times$ &\underline{3.58}$\times$&\underline{3.15}$\times$&\underline{3.01}$\times$ \\

\multicolumn{1}{c|}{}&V-13B & 2.04$\times$ & 2.14$\times$ &2.06$\times$ &2.27$\times$&2.07$\times$&1.98$\times$ &2.05$\times$&1.93$\times$&1.89$\times$ &2.70$\times$&2.24$\times$&2.22$\times$ \\
\midrule

\multicolumn{1}{c|}{\multirow{3}{*}{DRAGIN+IDR$_2$}}& L2-7B & \underline{3.34}$\times$ & 2.77$\times$ &2.52$\times$ &\underline{4.09}$\times$&2.08$\times$&\underline{2.09}$\times$ &\underline{\textbf{3.39}}$\times$&\underline{1.98}$\times$&\underline{\textbf{1.95}}$\times$ &\underline{4.49}$\times$&2.58$\times$&2.58$\times$ \\

\multicolumn{1}{c|}{}& L2-13B & 3.25$\times$ & \underline{\textbf{3.21}}$\times$ &\underline{\textbf{2.61}}$\times$ &3.97$\times$&\underline{2.15}$\times$&2.01$\times$ &3.15$\times$&1.98$\times$&1.87$\times$ &4.25$\times$&\underline{2.86}$\times$&\underline{2.65}$\times$ \\

\multicolumn{1}{c|}{}&V-13B & \textbf{3.57}$\times$ & 2.81$\times$ &2.55$\times$ &\textbf{4.14}$\times$&2.29$\times$&2.15$\times$ &3.25$\times$&1.88$\times$&1.82$\times$ &\textbf{4.72}$\times$&\textbf{4.00}$\times$&\textbf{3.53}$\times$ \\
\bottomrule
\end{tabular}
}
\caption{The average speedup ratios of IDR$_2$ across different methods, models, and datasets. Pref. (prefilling) demonstrates acceleration achieved through CICS and IDGR. Deco. (decoding) shows speedup from IGPG. E2E (end-to-end) reflects the overall A-RAG acceleration, encompassing retrieval.
($n$ = 3, 1$\times$ = baseline speed)}
\label{tab:sppedup_topk3}
\end{table*}

\begin{table*}[h]
\centering
\resizebox{0.85\textwidth}{!}{
\begin{tabular}{c|c|cc|cc|c|cc}
\toprule
\multicolumn{1}{c|}{\multirow{2}{*}{LLMs}}     & \multirow{2}{*}{Methods} & \multicolumn{2}{c|}{2WQA} &\multicolumn{2}{c|}{HQA} &\multicolumn{1}{c|}{SQA} &\multicolumn{2}{c}{IIRC}                                  \\ 
\cline{3-9} 
\multicolumn{1}{c|}{}                            &                          & \multicolumn{1}{c}{EM $\uparrow$}  & \multicolumn{1}{c|}{F1 $\uparrow$}&\multicolumn{1}{c}{EM $\uparrow$} & \multicolumn{1}{c|}{F1 $\uparrow$} &\multicolumn{1}{c|}{Acc. $\uparrow$} &\multicolumn{1}{c}{EM $\uparrow$} & \multicolumn{1}{c}{F1 $\uparrow$} \\ 
\midrule
\multirow{2}{*}{Llama2-7B} & DRAGIN$\dagger$ &22.5  &28.68&22.6&\textbf{33.02}&\textbf{65.10}&15.93&20.24                 \\
& DRAGIN+Ours&\textbf{25.4} & \textbf{33.17}&\textbf{22.9}&32.59&62.50&\textbf{16.04}&\textbf{20.24}\\
\midrule
\multirow{2}{*}{Llama2-13B}&DRAGIN$\dagger$&30.4&39.91&\textbf{31.6}&\textbf{42.60}&\textbf{66.10}&18.5&\textbf{22.59}                 \\
& DRAGIN+Ours &\textbf{34.4}&\textbf{41.50}&29.9&41.01&65.77&\textbf{18.55}&22.11\\
\midrule
\multirow{2}{*}{vicuna-13b}&DRAGIN$\dagger$&25.4&34.90&30.1&42.50&66.20&\textbf{23.90}&\textbf{28.11}                  \\
& DRAGIN+Ours &\textbf{28.9}&\textbf{36.77}&\textbf{31.7}&\textbf{42.58}&\textbf{68.00}&21.59&26.28\\
\bottomrule
\end{tabular}
}
\caption{The performance comparison on different datasets with different models. The $\dagger$ represents the reproduction.}
\label{tab:acc1}
\end{table*}

\subsection{Downstream Task Datasets}
We use the 2WikiMultihopQA \cite{ho2020constructing}, HotpotQA \cite{yang2018hotpotqa}, StrategyQA \cite{geva2021did} and IIRC \cite{ferguson2020iirc} as DRAGIN \cite{su2024dragin}.
The datasets span multi-hop question answering, common sense reasoning, and reading comprehension, covering different types of reasoning tasks. 

\subsection{Baselines}
We selected four representative A-RAG methods as baselines. These methods are characterized by two key aspects: retrieval timing and query construction. Retrieval timing determines when to perform retrieval, while query construction decides what content to use for retrieval. These features directly impact both the quality and efficiency of A-RAG workflows.
FL-RAG \cite{khandelwal2019generalization,borgeaud2022improving,ram2023context} retrieves every $z$ tokens using previous tokens as queries, FS-RAG \cite{trivedi2023interleaving} performs retrieval after each sentence, FLARE \cite{jiang2023active} triggers retrieval for uncertain tokens, using the last generated sentence as the query, and DRAGIN \cite{su2024dragin} employs a dynamic approach based on content importance and uncertainty, utilizing attention distribution for query construction.

\subsection{Implementation Details}
For the retrieval modules, we select BM25 and follow DRAGIN \cite{su2024dragin}. Additionally, we also investigate replacing BM25 with SGPT \cite{muennighoff2022sgpt}, a dense retrieval method. Our external knowledge is based on Wikipedia articles, which are divided into passages containing 100 tokens each. 
For the 7B and 13B models, we use four Nvidia 3090 GPUs and H800 GPUs, respectively. $n$ is the number of retrieved documents. Our IDR$_2$ based on the PyTorch \cite{paszke2019pytorch}.

\subsection{Main Results}
Table \ref{tab:sppedup_topk3} shows the results of our approach applied to the different A-RAG methods, different models with scales, and different scenarios. Table \ref{tab:sppedup_topk3} shows the acceleration ratios over baseline methods.

\begin{figure*}[h]
  \centering
  \begin{subfigure}[b]{0.3\textwidth}
    \includegraphics[width=\textwidth]{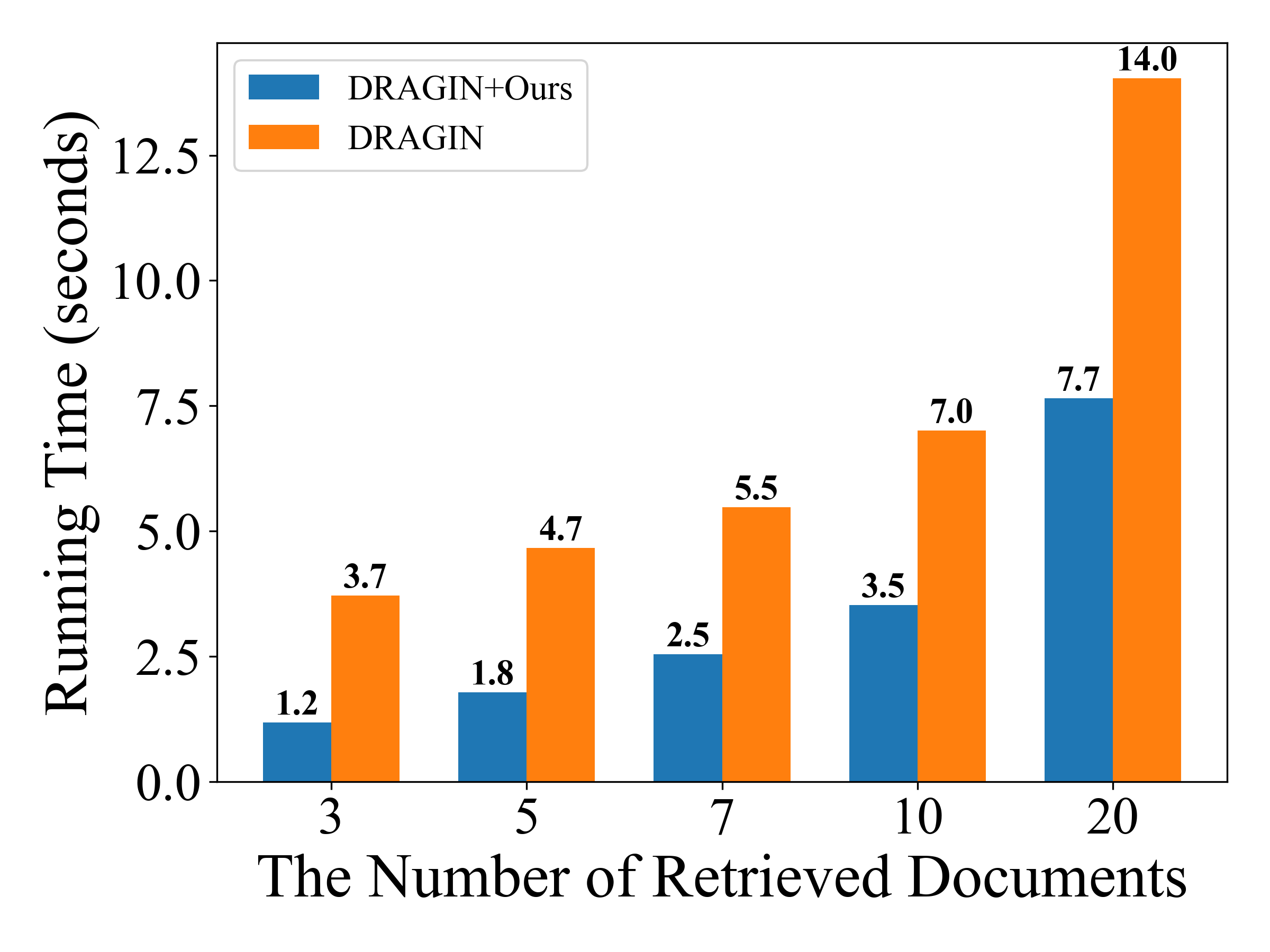}
    \caption{LLaMA2-7B prefilling cost.}
    \label{fig:subfig1.1}
  \end{subfigure}
  \hspace{2mm}
  \begin{subfigure}[b]{0.3\textwidth}
    \includegraphics[width=\textwidth]{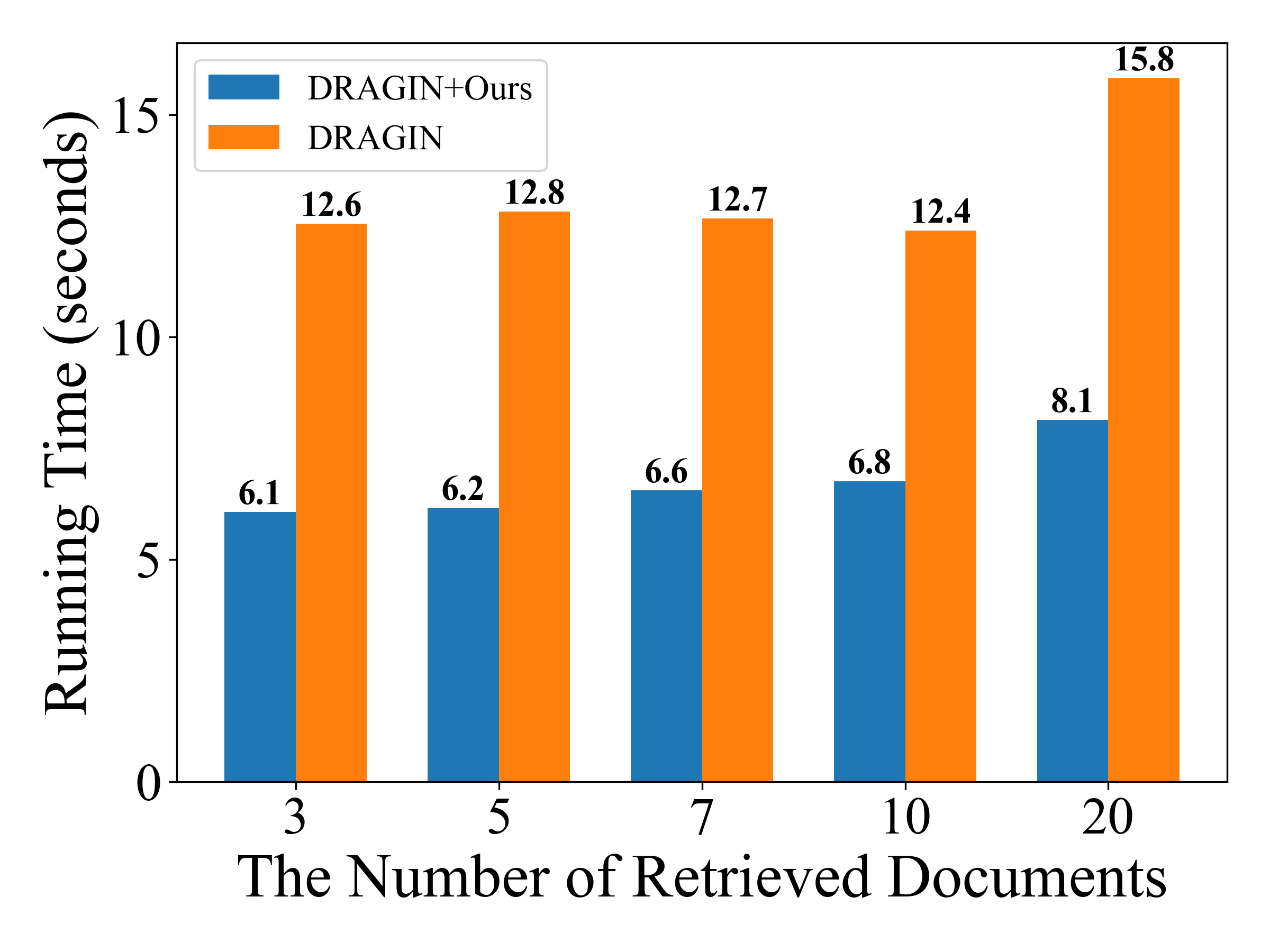}
    \caption{LLaMA2-7B decoding cost.}
    \label{fig:subfig1.2}
  \end{subfigure}
  \hspace{2mm}
  \begin{subfigure}[b]{0.3\textwidth}
    \includegraphics[width=\textwidth]{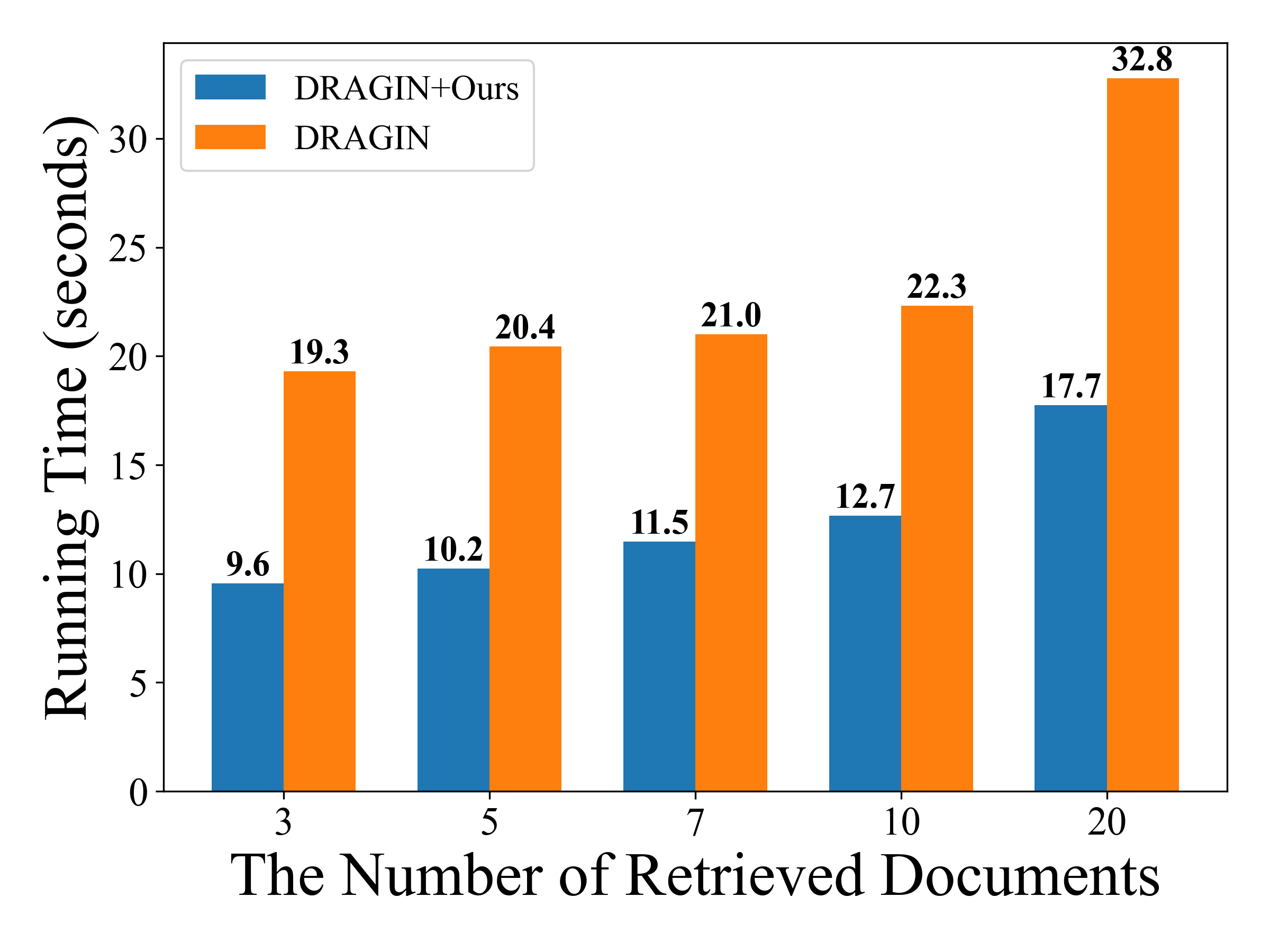}
    \caption{LLaMA2-7B end-to-end cost.}
    \label{fig:subfig1.3}
  \end{subfigure}
  \vspace{0.3em}
  
  \begin{subfigure}[b]{0.3\textwidth}
    \includegraphics[width=\textwidth]{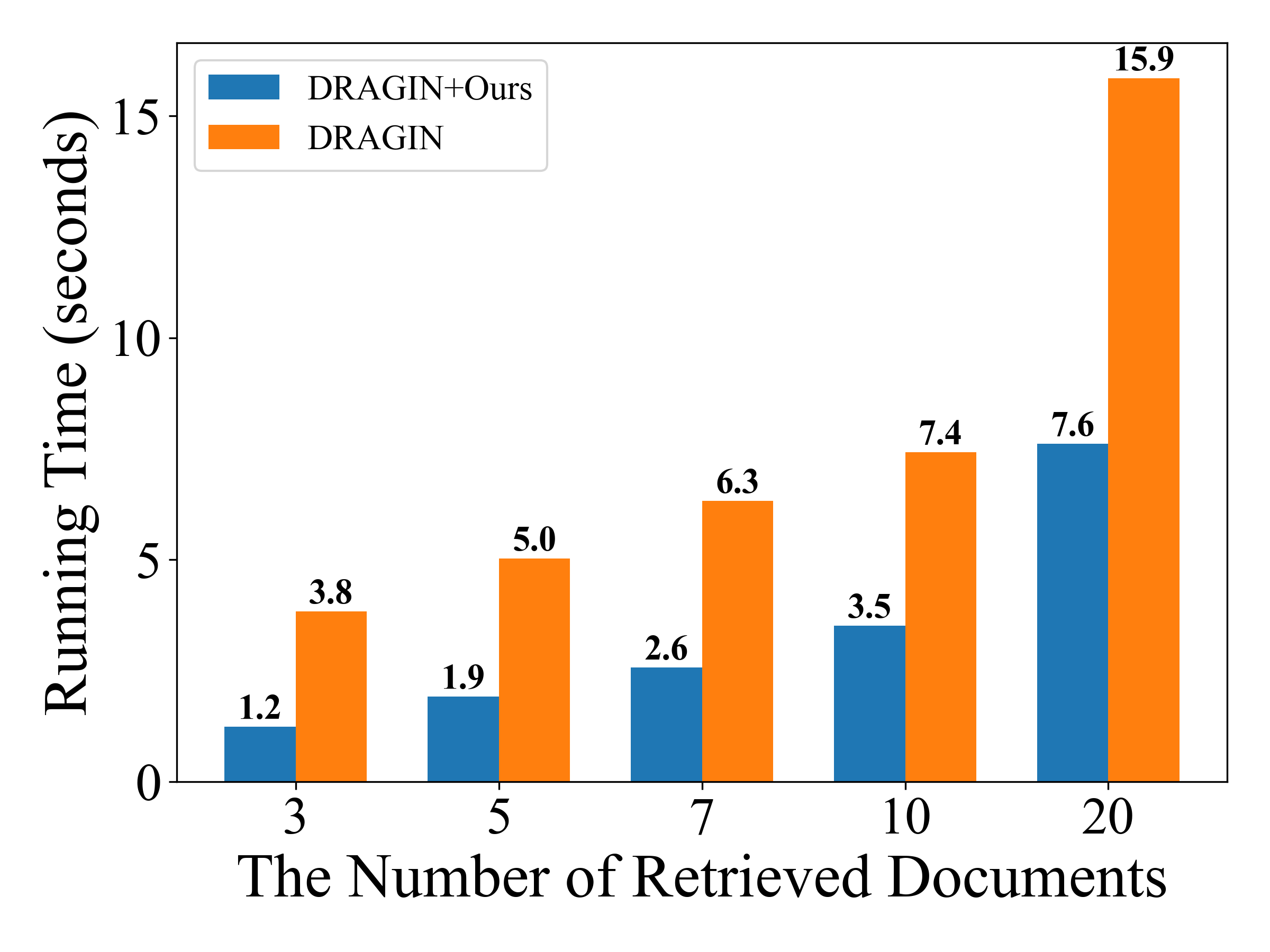}
    \caption{LLaMA2-13B prefilling cost.}
    \label{fig:subfig2.1}
  \end{subfigure}
  \hspace{2mm}
  \begin{subfigure}[b]{0.3\textwidth}
    \includegraphics[width=\textwidth]{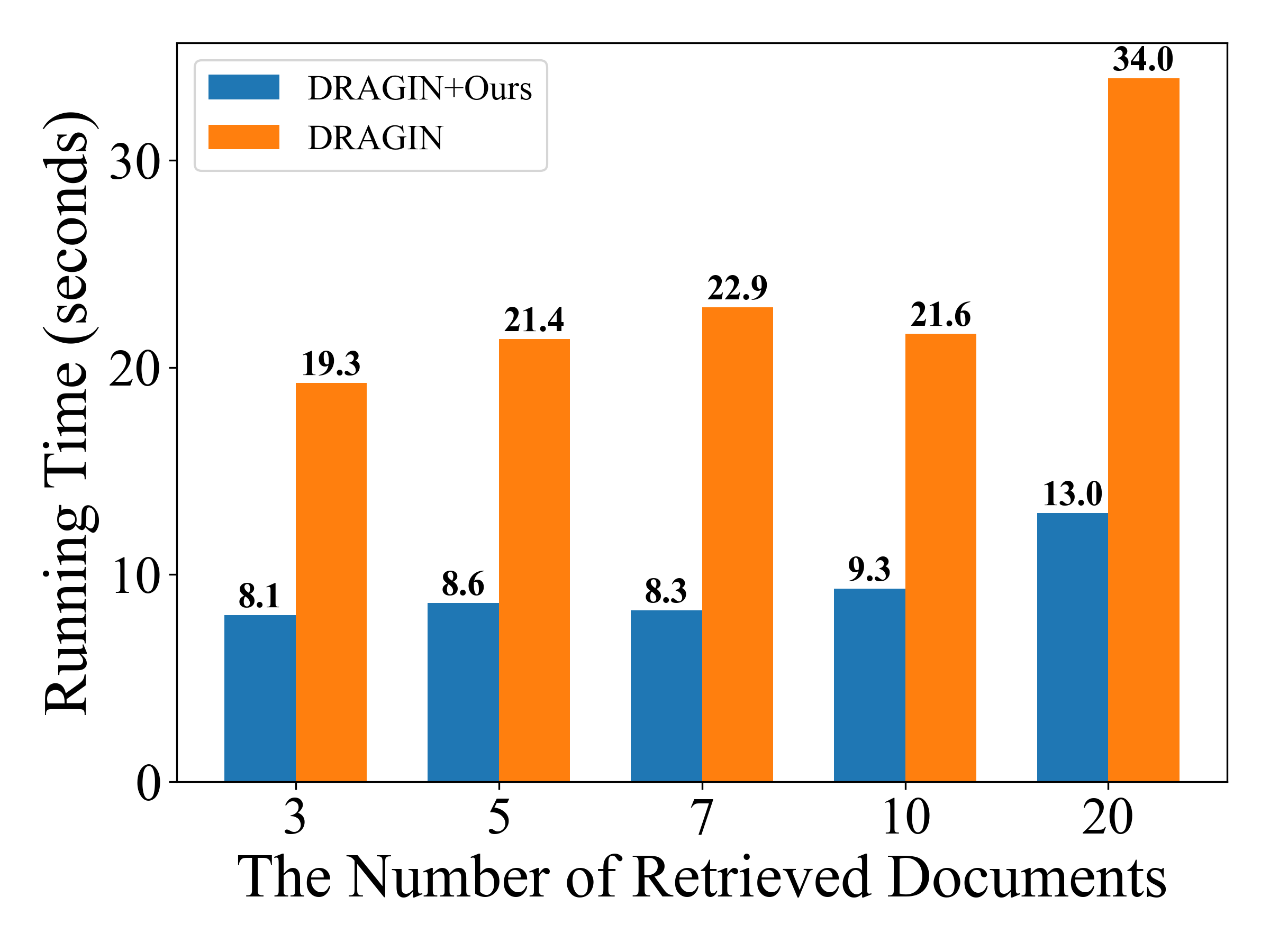}
    \caption{LLaMA2-13B decoding cost.}
    \label{fig:subfig2.2}
  \end{subfigure}
  \hspace{2mm}
  \begin{subfigure}[b]{0.3\textwidth}
    \includegraphics[width=\textwidth]{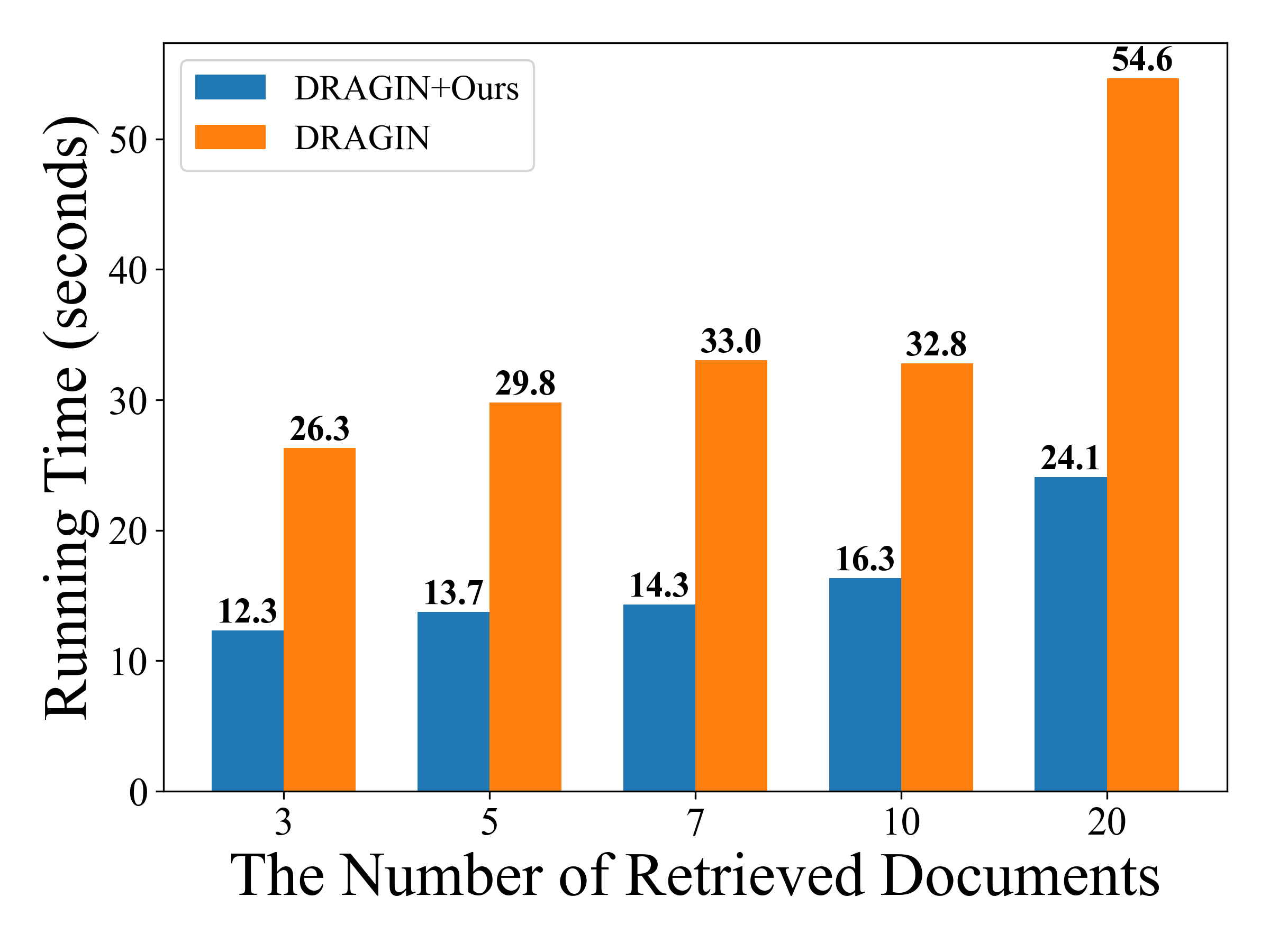}
    \caption{LLaMA2-13B end-to-end cost.}
    \label{fig:subfig2.3}
  \end{subfigure}
  \caption{The analysis of speedup for different numbers of retrieved documents.}
  \label{fig:diff_topk.1}
\end{figure*}

Overall, our method demonstrates end-to-end speed improvements ranging from 1.31$\times$ to 3.53$\times$. During the prefilling phase, we observe acceleration factors from 1.75$\times$ to 4.72$\times$. These results substantiate the efficacy of our proposed CICS and IDGR mechanisms in effectively eliminating redundant representations. For the decoding phase, the IGPG technique achieves speedup ratios from 1.49$\times$ to 4.00$\times$ through the reduction of autoregressive inference iterations.
A comparative analysis between the LLaMA2-7B and 13B architectures (underlining indicates better results) shows slightly better results for the 7B variant. We consider that this effect stems from the fact that smaller models require less additional overhead to store and load representations. Of the various A-RAG methods, the most significant performance improvement (4.72$\times$ for prefilling) is achieved when applying IDR$_2$ to DRAGIN. We attribute this to the query refinement mechanism of DRAGIN which improves query similarity, a property that effectively synergizes with our representation reduction approach.

\begin{table}[!htbp]
\centering
\resizebox{0.48\textwidth}{!}{
\begin{tabular}{c|c|ccc}
\toprule
{Models} & Methods & Pref. (s)$\downarrow$ & Deco. (s)$\downarrow$ & E2E (s)$\downarrow$\\
\midrule
\multicolumn{1}{c|}{\multirow{2}{*}{L2-7B}} & DR. & 3.71 & 12.55 & 19.31\\
\cline{2-5}
\multicolumn{1}{c|}{} & DR.+Ours & \textbf{1.18} & \textbf{6.07} & \textbf{9.56} \\
\midrule
\multicolumn{1}{c|}{\multirow{2}{*}{L2-13B}} & DR. & 3.84 & 19.26 & 26.31 \\
\cline{2-5}
\multicolumn{1}{c|}{} & DR.+Ours & \textbf{1.24} & \textbf{8.06 }& \textbf{12.34 }\\
\midrule
\midrule
\multicolumn{1}{c|}{\multirow{2}{*}{V-13b}} & DR. & 4.34 & 25.03 & 33.47 \\
\cline{2-5}
\multicolumn{1}{c|}{} & DR.+Ours & \textbf{1.29} & \textbf{10.49} & \textbf{15.18} \\
\bottomrule
\end{tabular}
}
\caption{The runtime of IDR$_2$ on 2WikiMultihopQA dataset ($n$ = 3), DR. denotes DRAGIN \cite{su2024dragin}. 
}
\label{tab:runtime_topk3}
\end{table}

The specific runtime of the different phases in the generation process are exhibited in Table \ref{tab:runtime_topk3}. Our experiments reveal that even on high-end NVIDIA H800 GPUs, 13B-parameter models require 26.31-33.47 seconds to process a single request. The proposed method significantly reduces the overall latency to 9.56-15.18 seconds, demonstrating substantial performance improvements and enhanced practicality for real-world deployment.

Table \ref{tab:acc1} shows the performance of applying the IDR$_2$ to the DRAGIN method. The experimental results show that our approach effectively preserves the original performance while exhibiting strong adaptability and generalization across different models and tasks. Notably, our method also carries some enhancements on 2WikiMultiHopQA. 

\subsection{Ablation studies}
\subsubsection{Impact of Retrieval Size}
We analyze the impact of varying numbers of retrieved documents on A-RAG, with results visualized in Figure \ref{fig:diff_topk.1}. As the document count increases, the context length grows monotonically, resulting in progressively higher prefilling overhead proportion. Our approach maintains consistent speedup advantages across both prefilling and decoding.
\begin{figure*}[h]
    \centering
    \includegraphics[width=0.95\linewidth]{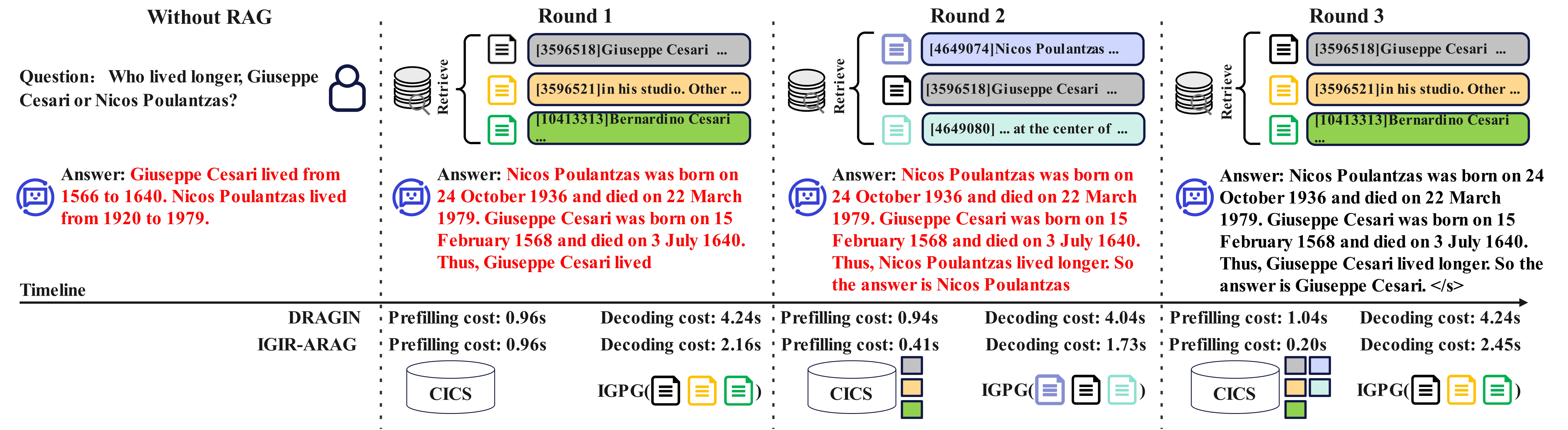}
    \caption{The detailed process of A-RAG based on a specific example
    , with LLaMA-13B.}
    \label{fig:case_study.1}
\end{figure*}

\subsubsection{Impact of Different Retrievers}
\begin{table}[h]
    \centering
    \begin{tabular}{c|c|c|c}
    \toprule
        Method & Retriever & EM $\uparrow$ & E2E $\uparrow$\\
        \midrule
        DRAGIN & BM25& 30.4 & 1$\times$ \\
        DRAGIN + IDR$_2$& BM25 & 34.4& 2.61$\times$\\
        \midrule
      DRAGIN & SGPT &     27.3 & 1$\times$ \\
         DRAGIN +IDR$_2$& SGPT & 29.6& 2.73$\times$\\
        \bottomrule

    \end{tabular}
    \caption{Analysis of different retrievers on the 2WikiMultihopQA with LLaMA2-13B ($n=3$).}
    \label{tab:retriever}
\end{table}
Table \ref{tab:retriever} demonstrates the effectiveness of IDR$_2$ across different retrievers. Substituting the retriever with SGPT results in an EM score of 29.6 and a speedup ratio of 2.73 $\times$. The consistent performance and acceleration above 2.6$\times$ across both frequency-based and semantic retrievers confirms the adaptability of IDR$_2$. 

\section{Analysis}
\subsection{Do these modules impact performance?}
Table \ref{tab: module_ablation} shows how different modules affect the generation quality. The redundant information in the CICS module impacts the generation quality of LLM. This is evident in the performance of LLaMA2-7B, where EM dropped from 22.0 to 20.3, with LLaMA2-13B showing similar decreases. The introduction of IDGR not only alleviates the adverse effects of redundant information but also brings additional performance gains. The LLaMA2-7B achieves EM scores of 25.4, and LLaMA2-13B reaches 34.4. These results confirm that IDGR successfully maintains inference quality.

\subsection{Specific Case Studies}
Figure \ref{fig:case_study.1} shows three iterative rounds. In \textbf{Round 1} (empty CICS), baseline prefilling takes 0.96s. IGPG reduces decoding time from 4.24s to 2.16s. After hallucination detection, the process enters \textbf{Round 2}. IDR$_2$ reuses the cached representations of \#359518 in CICS, saving 0.53s (prefilling). During the decoding process, IGPG builds fragment models, saving 2.31 seconds. For example, \textit{Nicos Poulantzas} requires six token-level autoregression: \textit{\_N}, \textit{icos}, \textit{\_P}, \textit{oul}, \textit{ant}, \textit{zas}. After applying IGPG, only one autoregression is needed after the prefix "Nicos," is generated, achieving 4$\times$ local acceleration. Similarly, dates like \textit{October 1979}, \textit{September 1936}, \textit{February 1568} achieve 4$\times$ acceleration by retrieving corresponding fragments from IGPG using month information.
\begin{table}[!htbp]
\centering
\resizebox{0.48\textwidth}{!}{
\begin{tabular}{ccc|cc|cc}
\toprule
\multirow{2}{*}{CICS} & \multirow{2}{*}{IGPG} & \multirow{2}{*}{IDGR} & \multicolumn{2}{c|}{LLaMA2-7B} & \multicolumn{2}{c}{LLaMA2-13B} \\
\cline{4-7}
& & & EM $\uparrow$ & F1 $\uparrow$& EM $\uparrow$& F1 $\uparrow$\\
\midrule
- & - & - & 22.5 & 28.68 & 30.4 & 39.91\\
\midrule
\checkmark & - & - & 20.3 & 26.88 & 28.0 & 37.49\\
- &\checkmark&- & 22.4 &28.51 & 30.4 &40.02 \\
\checkmark & \checkmark & - & 20.2 & 26.78 & 28.3 & 37.62\\
\checkmark & \checkmark & \checkmark & \textbf{25.4} & \textbf{33.17} & \textbf{34.4} & \textbf{41.50} \\
\bottomrule
\end{tabular}
}
\caption{Analysis of the impact of different modules on 2WikiMultihopQA based on DRAGIN.}
\label{tab: module_ablation}
\end{table}
At \textbf{Round 3}, LLM further confirms \textit{Giuseppe Cesari's} lifespan and generates the right answer. With all representations already in CICS, the prefilling time reduces by 0.804 seconds (about 80\% of DRAGIN's original time), achieving 5× speedup. Total savings reach 1.334s (prefilling) and 6.16s (decoding), summing to 7.494s (48.47\% reduction) over DRAGIN's 15.46s, achieving 1.95× end-to-end acceleration. 
\section{Conclusion}
This paper presents IDR$_2$, a model-agnostic approach that significantly improves the efficiency of A-RAG by addressing redundant representation processing across multiple iterations. To prevent redundant computation, this paper introduces the CICS module for caching document representations. The IDGR module then guides the generation process to focus on crucial information through the instruction-following capability of LLMs. Furthermore, the IGPG module leverages the correlation between generated content and documents to enable parallel generation of existing content, reducing autoregressive rounds. Through extensive experiments, 
IDR$_2$ consistently demonstrates remarkable efficiency improvements, achieving up to 2.0$\times$ acceleration while maintaining generation quality. Our future work aims to further compress the KV cache while preserving its representation capabilities.

\section{Limitations}
We acknowledge the limitations of this paper. Although our method is a general approach, it requires the corresponding LLM to be open-source to apply the CICS and IGPG technologies in this paper. It should be noted that the method in this paper is not applicable to LLM APIs that only support text-based interfaces. Therefore, our future work also aims to develop more methods to overcome the limitations of similar scenarios.

\bibliography{custom}

\end{document}